# Deep Learning for Vanishing Point Detection Using an Inverse Gnomonic Projection

Florian Kluger[1], Hanno Ackermann[1], Michael Ying Yang[2], and Bodo Rosenhahn[1]

[1] Leibniz Universität Hannover
[2] University of Twente

**Abstract.** We present a novel approach for vanishing point detection from uncalibrated monocular images. In contrast to state-of-the-art, we make no a priori assumptions about the observed scene. Our method is based on a convolutional neural network (CNN) which does not use natural images, but a Gaussian sphere representation arising from an inverse gnomonic projection of lines detected in an image. This allows us to rely on synthetic data for training, eliminating the need for labelled images. Our method achieves competitive performance on three horizon estimation benchmark datasets. We further highlight some additional use cases for which our vanishing point detection algorithm can be used.

## 1 Introduction

Vanishing points (VPs) are strong cognitive cues for the human visual perception, as they provide characteristic information about the geometry of a scene, and are used as a feature for relative depth and height estimation [23]. Their detection is a fundamental problem in the field of computer vision, because it underpins various higher-level tasks, including camera calibration [12, 15, 25], 3D metrology [8], 3D scene structure analysis [13], as well as many others. A vanishing point arises from a set of parallel lines as their point of intersection, at an infinite location initially, and is uniquely defined by the lines' direction. Under a projective transformation, parallel lines in space may be transformed to converging lines on an image plane, thus leading to a finite intersection point. The detection of VPs in perspective images is therefore a search for converging lines and their intersections, which is difficult in the presence of noise, spurious line segments, near-parallel imaged lines, and intersections of non-converging lines. These reasons make vanishing point detection a hard problem. Consequently, it has not been addressed often in the past years.

### 1.1 Related Work

Since the seminal work of Barnard [4], various methods designed to tackle this problem have been proposed. Some of them [15, 20, 22, 25] rely on the Manhattan-world assumption [7], which means that only three mutually orthogonal vanishing directions exist in a scene, as is reasonably common in urban scenes where buildings are aligned on a rectangular grid. Others [3, 18, 21, 27, 28] rely on the less rigid Atlanta-world assumption [21], which allows multiple non-orthogonal vanishing directions that are connected by a common horizon line, and are all orthogonal to a single zenith. Few works [1, 2, 24] – including ours – make no such assumptions. Most methods are based on oriented elements – either line segments [18, 27, 25, 15, 20] or edges [3, 22, 21] – from which VPs are estimated, usually by grouping the oriented elements into clusters [18, 27, 25, 22, 15], or by fitting a more comprehensive model [2, 3, 24]. It is common to refine the thereby detected vanishing points in an iterative process, such as the Expectation Maximisation (EM) algorithm [28, 27, 25, 22, 15] which we are utilising as well.
Ever since the AlexNet by Krizhevsky et al. [16] succeeded in the 2012 ImageNet competition, convolutional neural networks (CNNs) [17] have become a popular tool for computer vision tasks, as they perform exceedingly well in a variety of settings. Borji [6] recently demonstrated a CNN based approach for vanishing point detection; however, it only detects up to one horizontal vanishing point. The approach of Zhai et al. [28] is much more comprehensive. It is based on a CNN which extracts prior information about the horizon line location from an image, and currently achieves the best state-of-the-art performance on horizon line detection benchmarks commonly used to evaluate vanishing point detection algorithms. Unlike other methods, their approach begins by selecting horizon line candidates first, and then jointly scores horizon candidates and horizontal VP candidates, which are eventually refined in an EM-like process. As it is based on horizon lines, their approach is inherently limited to Atlanta-world scenes.

**Contributions** In this work, we propose a more generalised approach using a CNN which does not operate on natural images, but on a more abstract presentation of the scene based on the Gaussian sphere representation of points and lines [4]. It is identical to an inverse gnomonic projection, which – very similar to an inverse stereographic projection – is a mapping that transforms the unbounded image plane onto a bounded space, thus making vanishing points far from the image centre easier to handle. While this necessitates an additional preprocessing step, it allows us to train the CNN solely using synthetic data, which we generate in a very straightforward manner, thus eliminating the need for labelled real-world data. The use of the CNN is motivated by the fact that spurious, yet significant VP candidates can occur (cf. Fig. 4), thus approaches based on voting are prone to fail. The advantage of using a CNN over, for instance, a support vector machine is that discriminative features are automatically learned. Our CNN is able to directly estimate VP candidates on the Gaussian sphere, which are then refined with an EM-like algorithm. We furthermore devised an improved line weighting scheme for the EM process, which imposes a spatial consistency prior over the line-to-vanishing-point associations in order to become more robust in the presence of noise and clutter. This is motivated by the fact that spurious lines, for instance caused by plants or shadows, are often spatially correlated. Combined with a line segment extractor as a preprocessing step, our method allows vanishing point estimation from real-world images with competitive accuracy.

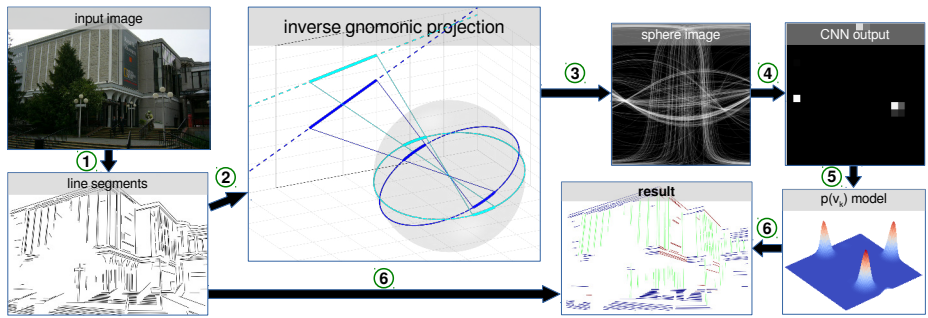

Fig. 1: Algorithm pipeline: 1) extract line segments; 2) map lines onto Gaussian sphere with inverse gnomonic projection (Sec. 2.1); 3) render image of half-sphere surface (Sec. 2.1); 4) compute CNN (Sec. 2.2) forward pass; 5) estimate a mixture of Gaussian distribution after CNN output (Sec. 2.4); 6) compute refined vanishing points and visualise line association (Sec. 2.4).

## 2   Approach

Our approach consists of the following stages: First, line segments are extracted from the input image using the LSD line detector [11]. The lines are then mapped onto the Gaussian sphere, and its image is rendered (Sec. 2.1). This image is used as the input for a CNN (Sec. 2.2) which we trained solely on synthetic data (Sec. 2.3). This CNN then provides a coarse prediction of possible VP locations, which are ultimately refined in an Expectation Maximisation based process (Sec. 2.4).

### 2.1   Parametrisation

In order to deal with infinite vanishing points, it is reasonable to transform the unbounded image plane onto a bounded space, such as the Gaussian sphere representation, which based on an inverse gnomonic projection of homogeneous points $\mathbf{p} = (p_1, p_2, p_3)^T$ and lines in normal form $\mathbf{l} = (l_1, l_2, l_3)^T$, as described by Barnard [4]:

$$\frac{\mathbf{p}}{\|\mathbf{p}\|_2} = (\sin\alpha\cos\beta, \sin\beta, \cos\alpha\cos\beta)^T \quad (1)$$

$$\beta(\alpha, \mathbf{l}) = \tan^{-1}\left(\frac{-l_1\sin\alpha - l_3\cos\alpha}{l_2}\right) \quad (2)$$

The lines are projected from the image plane at a fixed distance onto the unit sphere at origin. A square image of the sphere's front half surface is rendered, so that the lines appear as opaque curves, and the image's x,y-coordinates correspond to azimuth and elevation $(\alpha, \beta)$ on the sphere. This *sphere image* (cf. Fig. 1) is later used as an input for the CNN. The vanishing points are likewise parametrised in the $\alpha, \beta$-space, and are then quantised into bins on a regular $N \times N$ grid, so that the occurrence of a vanishing point within those bins can be treated as a multi-label classification task.

*Normalisation:* As actual images can be of various sizes, image coordinates are normalized to fit within a $(-1, 1) \times (-1, 1)$ border by applying the following, aspect ratio preserving, transformation:

$$\mathsf{H}_{norm} = \frac{1}{s} \begin{pmatrix} 2 & 0 & -w \\ 0 & 2 & -h \\ 0 & 0 & s \end{pmatrix} \quad (3)$$

with $h$ and $w$ being the image's height and width, respectively, and $s = \max(w, h)$.

### 2.2   Network Architecture

We used the popular AlexNet [16] as a basic architecture for our approach. This network consists of five convolutional layers – some of them extended by max pooling, local response normalisation, or ReLU layers – followed by three fully connected (FC) layers. Originally, its final layer has 1000 output nodes, to which a softmax function is applied, and a multinomial logistic loss function is used for training, as is common for *one-of-many* classification tasks.

While a regression approach may seem well suited for a task such as vanishing point detection from line segments, training CNNs for regression tasks is notoriously difficult. We therefore decided to reformulate it as a multi-label classification task by partitioning the surface of the Gaussian sphere into $N \times N$ patches, as described in Sec. 2.1, and assigning distinct class labels to each patch. In order to suit our task, we modified the last FC layer of the AlexNet to contain $N^2$ output nodes, and replaced the softmax with a sigmoid function. For training, we use a cross entropy loss function, which is well suited for multi-label classification tasks like this. The output of the network is a likelihood image of possible vanishing points in the given scene.

### 2.3   Training Data

Since suitable training data for vanishing point detection tasks is scarce, and compiling annotated data on our own would have been very laborious and time consuming, yet large amounts of training data are needed to obtain good results with deep learning, we decided to solely rely on a synthetically created dataset. Since our approach does not actually need natural images as an input, but relies on line segments only, we can create such synthetic data without much effort. To create a set of line segments as a piece of data for training, we proceed as follows: First, the number of vanishing directions $K_d \in [1, 6]$ is chosen. The first three (or less) directions are then chosen randomly, but with the condition that they must be mutually orthogonal. Additional directions are set as a linear combination of two randomly chosen, previously set directions, thus vanishing directions 4 to 6 do not form an orthogonal system. For each direction, a varying number of line clusters is placed in 3D space. Each cluster consists of a varying number of parallel or collinear line segments in close proximity. Additionally, some

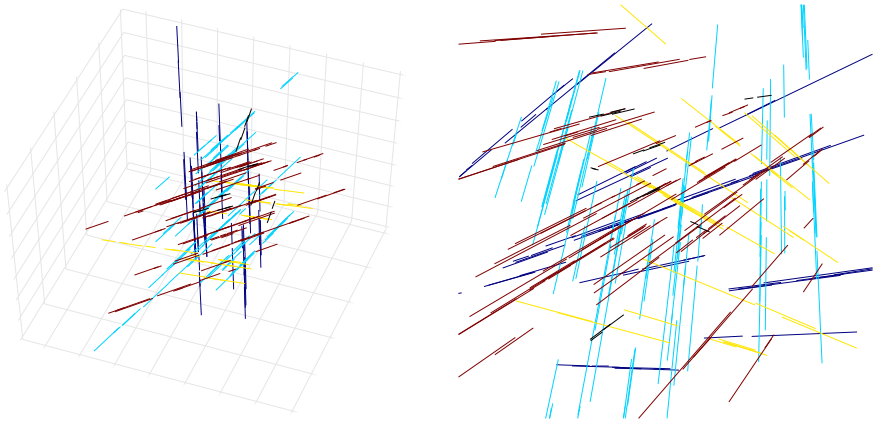

Fig. 2: Example from the synthetic training dataset (Sec. 2.3) with four vanishing directions. Different line colours denote different directions, with outlier lines shown in black. *Left:* 3D line segment plot. *Right:* 2D projection.

*outlier* line segments which are not aligned with any of the vanishing directions are interspersed. This 3D scene is then projected into 2D using a virtual pinhole camera with randomly chosen rotation, translation and focal length. Either uniform or Gaussian noise is added to the resulting 2D line segments, with its strength varying from example to example. These line segments are then finally cropped to fit within a $(-1, 1) \times (-1, 1)$ border. One example is shown in Fig. 2. Using this procedure, we create 96,000 examples for each number of vanishing directions, resulting in a dataset of 576,000 training examples. Each line segment is then converted into a line in normal form, and the true vanishing point for each vanishing direction is computed, so that every datum can be parametrised for CNN training as described in Sec. 2.1.

### 2.4   Vanishing Point Refinement

As the response of the CNN is rather coarse, a post-processing step is needed to determine the exact vanishing point locations. We decided to utilise a variant of the Expectation Maximisation (EM) algorithm, based on the method described by Košecká and Zhang [15] with additional modifications:

*E-step:* An affinity measure $w_{ik}$ between a line segment $l_i$ – or its corresponding homogeneous line $\mathbf{l}_i$ – and a vanishing point candidate $\mathbf{v}_k$ is calculated based on the posterior distribution:

$$w_{ik} \propto p(\mathbf{v}_k|\mathbf{l}_i) = \frac{p(\mathbf{l}_i|\mathbf{v}_k)\ p(\mathbf{v}_k)}{p(\mathbf{l}_i)} \tag{4}$$

with $p(\mathbf{l}_i) = \sum_k p(\mathbf{v}_k) p(\mathbf{l}_i|\mathbf{v}_k)$. We assume a likelihood modelled by:

$$p(\mathbf{l}_i|\mathbf{v}_k) \propto \exp\left(\frac{-d_{ik}^2}{2\sigma_k^2}\right) \tag{5}$$

with $d_{ik}$ being a consistency measure between $l_i$ and $\mathbf{v}_k$.

*M-Step:* New vanishing point estimates are obtained by solving the following least-squares problem:

$$J(\mathbf{v}_k) = \min_{\mathbf{v}_k} \sum_i w_{ik} d_{ik}^2 \quad . \tag{6}$$

**Modifications** As in [15], we measure the distance $d_{ik}^{(1)} = \mathbf{l}_i^T \mathbf{v}_k$ on the Gaussian sphere to solve (6), but use an angle-based consistency measure, similar to the suggestions of [9, 20], to compute (5), as this yields better accuracy. With $\mathbf{m}_i$ being the midpoint of $\mathrm{l}_i$ and $\mathbf{m}_i \times \mathbf{v}_k$ denoting a cross-product, we define:

$$d_{ik}^{(2)} = 1 - \cos(\angle(\mathbf{l}_i, \mathbf{m}_i \times \mathbf{v}_k)) \tag{7}$$

Departing from [15], we utilise the output of our CNN to estimate the prior $p(\mathbf{v}_k)$ and to initialise the VP candidates, and furthermore propose a modified affinity measure $w_{ik}$ to consider the spatial structure of line segments.

*Vanishing point prior:* We treat the output of the CNN as an approximation of the true probability density distribution for $p(\mathbf{v}_k)$ in the $(\alpha, \beta)$-space, which we model as a mixture of Gaussians with $N^2$ components of standard deviation $\sigma_{prior}$. Each component is located at the centre of the corresponding patch on the Gaussian sphere and weighted proportionally to its CNN response. This is illustrated in Fig. 1.

*Initialisation:* First, the $K_{init}$ strongest local maxima of the CNN response are detected. Each of these corresponds to a patch on the Gaussian sphere image (cf. Fig. 1). Then, the global maximum within such a patch is detected and its position on the sphere converted back to euclidean coordinates, which yields an initial vanishing point candidate.

*Affinity measure:* Originally, $w_{ik} = p(\mathbf{v}_k | \mathbf{l}_i)$ was used in [15] as an affinity measure. As it does not take the spatial structure of line segments into account, we devised a modified affinity measure based on the following assumptions: *1.* Line segments with similar orientation in close proximity likely belong to the same vanishing point. *2.* Line segments that lie within a neighbourhood of similarly oriented lines less likely originate from noise. Based on this intuition, we devised a similarity measure $S_{ij}$ between two line segments $\mathrm{l}_i, \mathrm{l}_j$:

$$S_{ij} = \cos(\phi_{ij}) \exp\left(\frac{-d_l(\mathrm{l}_i, \mathrm{l}_j)^2}{\sigma_l^2}\right) \tag{8}$$

with $\phi_{ij} = \min\left(\max\left(k_\phi \cdot \angle(\mathbf{l}_i, \mathbf{l}_j), -\frac{\pi}{2}\right), \frac{\pi}{2}\right)$, and $d_l(l_i, l_j)$ being the shortest distance between the two line segments. Using this similarity measure, we enforce a prior on $w_{ik}$ in order to achieve higher spatial consistency between line segments w.r.t. their vanishing point associations. We further assign a higher relative weight to those line segments which, according to the similarity measure, appear to lie in a neighbourhood of other, similar line segments, assuming that this indicates a regular structure as opposed to noise.

*Split and merge:* In some cases, a detection would occur at a spot within the image's borders which is not a true vanishing point, but merely a point of coincidental intersection of lines. In order to counteract this, we devised a *split-and-merge* technique which is applied once every $f_s$ iterations of the EM process: First, a vanishing point within the image whose associated line segments have the highest standard deviation w.r.t. their angle is selected. Then, these line segments are split into two clusters based on their angle, from which two new vanishing points are calculated, replacing the old one. If one resulting vanishing point is too close to another, they will be merged together afterwards.

### 2.5  Horizon Line and Orthogonal Vanishing Point Estimation

As is customary, we used the horizon detection error metric to compare our approach to previous methods. We devised an algorithm that estimates three supposedly orthogonal vanishing points and a horizon line, given a set of previously determined vanishing points.

First, we select the $N_{vp}$ most significant vanishing points – where significance is measured by the number of lines $n_k$ associated with a vanishing point $\mathbf{v}_k$ – and consider every possible triplet $\mathcal{T}$. Any vanishing point with an elevation $|\beta_k| > \theta_z$ on the Gaussian sphere is considered as a zenith candidate $\mathbf{v}_z$. We then discard unreasonable solutions, e.g. those which would result in a horizon line slope $\phi_{hor} > \theta_{hor}$. We assume that the projection of the camera centre $\mathbf{c}$ coincides with the center of the image and calculate the angle $\phi_{hz,\mathcal{T}}$ between the tentative horizon line and the line $\mathbf{l}_{zc} = \mathbf{v}_z \times \mathbf{c}$, which ideally should be perpendicular [5]. Then we calculate a score value for each triplet:

$$s_\mathcal{T} = (1 - \cos(\phi_{hz,\mathcal{T}})) \cdot \sum_{i \in \mathcal{T}} n_i \qquad (9)$$

and select the triplet with the highest score. Finally, a horizon line $\mathbf{h}$ is calculated – under the condition that $\mathbf{h}$ and $\mathbf{l}_{zc}$ be perpendicular – by minimising:

$$J(\mathbf{h}) = \min_{\mathbf{h}} \sum_{i \in (\mathcal{T} \setminus zenith)} \frac{n_i}{\|\mathbf{v}_i - \mathbf{c}\|} (\mathbf{h}^T \mathbf{v}_i)^2 \qquad (10)$$

Table 1: Parameters of our method used for all experiments.

| Name | $N$ | $K_{init}$ | $\sigma_{prior}$ | $k_\phi$ | $N_{vp}$ | $\theta_z$ | $\theta_{hor}$ |
|---|---|---|---|---|---|---|---|
| Section | 2.1, 2.2 | 2.4 | 2.4 | 2.4 | 2.5 | 2.5 | 2.5 |
| Value | 20 | 25 | $\frac{\pi}{1.282N}$ | 9 | 20 | $\frac{\pi}{4}$ | $\frac{\pi}{6}$ |

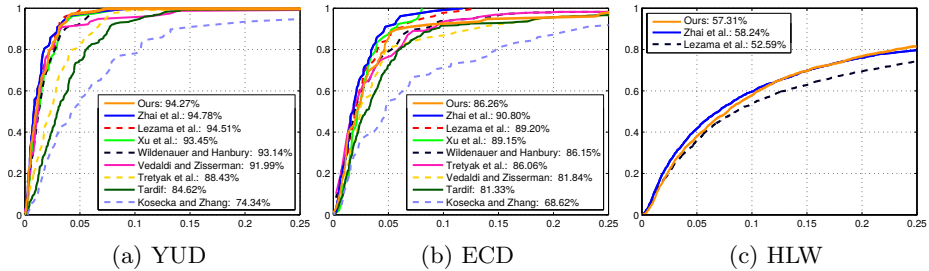

(a) YUD                (b) ECD                (c) HLW

Fig. 3: Cumulative histograms of the horizon detection error. The horizon error is represented on the x-axis, while the y-axis represents the fraction of images with less than the corresponding error.

## 3   Experiments

We implemented and trained our CNN with the Caffe [14] framework and used an existing C++ library [11] for line detection. All other pre- and post-processing steps were implemented in Python, making use of the Numpy and Scikit-learn [19] packages. The parameters in Tab. 1 were used for all experiments. On an Intel Core i7-3770K CPU, our implementation takes 45 seconds on average to compute the result for a 640x480 pixel image. The majority of this time – almost 95% – is needed for the EM based refinement step.

### 3.1   Horizon Estimation

For a quantitative evaluation of our method, we computed the horizon detection error on two benchmark datasets that were commonly used to assess the performance of vanishing point detection in previous works [3, 18, 24, 25, 27, 28], as well as a third, more recent dataset additionally used for evaluation in [28]. The horizon detection error is defined as the maximum distance between the detected and the true horizon line, relative to the image's height.

The *York Urban Dataset* (YUD) [9] contains 102 images of indoor and urban outdoor scenes, and three vanishing points corresponding to orthogonal directions are given as ground-truth for each scene. Generally, these scenes fulfil the Manhattan-world assumption, though our method does not take advantage of that. Fig. 3a shows the cumulative horizon error histogram and the area under the curve (AUC) as a performance measure, comparing our approach to competing methods. We achieve a competitive AUC of 94.27%, compared to 94.78% of the current best state-of-the-art method [28]. In contrast to [28], in which only the horizon line estimation is evaluated, we are able to identify the three orthogonal vanishing directions with an accuracy of 99.13% within a margin of error of five degrees.

The *Eurasian Cities Dataset* (ECD) [3] contains 103 urban outdoor scenes which generally do not satisfy the Manhattan-world assumption, but often contain multiple groups of orthogonal directions, and are therefore more challenging compared to the YUD. The horizon line and a varying number of vanishing points are given as ground-truth for each scene. On this dataset, we achieve an AUC of 86.26%. Fig. 3b gives a comparison to other state-of-the-art methods.

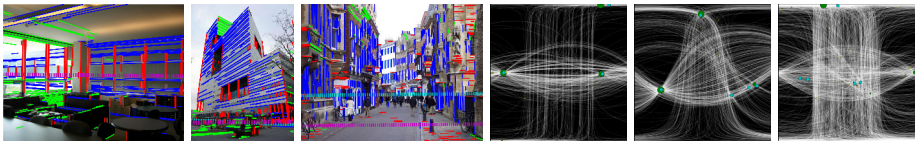

Fig. 4: First 3 images show line segments associated with the three VPs used for horizon estimation (red, green, blue), estimated horizon (magenta) and ground truth horizon (cyan). Images 4-6 show the corresponding sphere images with most significant (green) and other detected (yellow) VPs, and ground truth (cyan). The 1st and 4th images show the best case example, the 2nd and 5th an average case example, the 3rd and 6th a failure case.

The *Horizon Lines in the Wild* (HLW) dataset [26] is a recent benchmark dataset which is significantly more challenging than both YUD and ECD, as many of its approximately 2000 test set images do not fulfil the Atlanta-world assumption. Here, our method achieves 57.31% AUC – slightly worse than [28] with 58.24%, but vastly better than [18] with 52.59%, see Fig. 3c.

Generally, our method appears to perform poorly when a large number of line segments near the horizon, large curved structures, or a very large number of noisy line segments are present. A representative failure case is shown by the 3rd and 6th images in Fig. 4.

### 3.2 Additional Applications

As camera systems are an essential source of data for autonomous vehicles and driver assistance systems upon which they base their actions, it is reasonable to extract as much useful information as possible from the images they capture. We want to illustrate that robustly estimated vanishing points are of great use for such applications.

In order to extract metrically correct measurements within a scene from camera images, knowledge of the cameras intrinsic parameters $\mathsf{K}$ is required. While they can be acquired by calibration before deploying the camera in a vehicle, shock and vibration may alter the camera's internal alignment over time, resulting in a need for recalibration. Such a recalibration is possible by way of determining the image of the absolute conic $\omega = \mathsf{K}^{-T}\mathsf{K}^{-1}$ from three orthogonal vanishing points if zero skew and square pixels are assumed. A method that facilitates this is outlined in [12], while a simplified version that assumes the camera's principle point to be known – but only needs two orthogonal vanishing points – is described in [25].

If the camera's intrinsic parameters are known, a homography $\mathsf{H} = \mathsf{KRK}^{-1}$, which is akin to a rotation of the camera with a 3D rotational matrix $\mathsf{R}$, can be computed. This can be exploited to align one or two vanishing directions with the canonical $x$, $y$ or $z$-axes in a way that results in a rectification of any plane which is aligned with said directions. Such a rectification can be used to extract relative measurements within a plane, e.g. for computing relative widths within a traffic lane (cf. Fig. 5a-b), or to project auxiliary information – such as street names or traffic signs – into a scene (cf. Fig. 5c) and display it to the driver, thus facilitating a form of visually appealing augmented reality without the need for explicit 3D reconstruction.

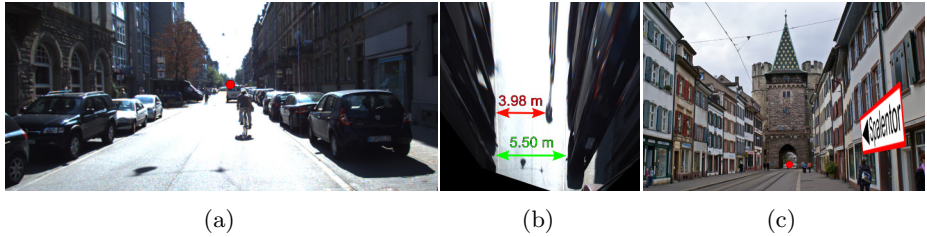

(a)　　　　　　　　　　　(b)　　　　　　　　　　(c)

Fig. 5: (a) Image from the KITTI [10] dataset, with a detected vanishing point (red dot) arising from the central perspective. (b) Rectified version of *(a)* after aligning the vanishing direction with the $y$-axis. The relative amount of space next to the cyclist can be measured easily. (c) Image[3] of the *Spalentor* in Basel with a virtual sign projected onto a wall after estimating $\mathsf{K}$ from detected vanishing points and aligning the central vanishing point with the $x$-axis.

## 4   Conclusion

We introduced a novel, deep learning based vanishing point detection method, which uses a CNN that operates on artificial images arising from a Gaussian sphere representation of lines and points using an inverse gnomonic projection. It is trained using synthetic data including noise and outliers exclusively, eliminating the need for labelled data. Despite not relying on either the Manhattan-world or Atlanta-world assumptions, which most related works do, it achieves competitive results on three benchmark datasets and good results in two further applications. Obviously, the capability of the trained CNN to handle different scenes depends on the training data. Since the proposed approach relies on synthetic data, it can be easily amended to represent different cases. The results on *Horizon Lines in the Wild* (HLW) demonstrates that the used training data is representative for difficult real images. Even more challenging scenarios, for instance no orthogonal VPs at all, can be approached by generating suitable training data and simply re-training the CNN.

**Acknowledgements:** This research was supported by German Research Foundation DFG within Priority Research Programme 1894 *Volunteered Geographic Information: Interpretation, Visualisation and Social Computing.*

---

[3] Source: https://www.flickr.com/photos/david-perez/4493470850